\documentclass{article}

\usepackage{arxiv}

\usepackage[utf8]{inputenc} 
\usepackage[T1]{fontenc}    
\usepackage{hyperref}       
\usepackage{url}            
\usepackage{booktabs}       
\usepackage{amsfonts}       
\usepackage{nicefrac}       
\usepackage{microtype}      
\usepackage{lipsum}
\usepackage{graphicx}
\graphicspath{ {./images/} }

\title{Benchmarking Vision-Language and Multimodal Large Language Models in Zero-shot and Few-shot Scenarios:
A study on Christian Iconography}

\author{
 Gianmarco Spinaci \\
  Department of Classical Philology and Italian Studies, University of Bologna, Italy\\
  Villa i Tatti, The Harvard University Center for Italian Renaissance Studies, Florence, Italy\\
  \texttt{gianmarco.spinaci2@unibo.it} \\
   \And
 Lukas Klic \\
  Villa i Tatti, The Harvard University Center for Italian Renaissance Studies, Florence, Italy\\
  \texttt{lklic@itatti.harvard.edu} \\
  \And
 Giovanni Colavizza \\
  Department of Classical Philology and Italian Studies, University of Bologna, Italy\\
  Department of Communication, University of Copenhagen, Denmark\\
  \texttt{giovanni.colavizza@unibo.it} \\
}

\begin{document}
\maketitle
\begin{abstract}
This study evaluates the capabilities of Multimodal Large Language Models (LLMs) and Vision Language Models (VLMs) in the task of single-label classification of Christian Iconography, focusing on their performance in zero-shot and few-shot settings across curated datasets. The goal was to assess whether general-purpose VLMs (\textit{CLIP} and \textit{SigLIP}) and LLMs, such as \textit{GPT-4o} and \textit{Gemini 2.5}, can interpret the Iconography, typically addressed by supervised classifiers, and evaluate their performance. Two research questions guided the analysis: \textbf{(RQ1)} How do multimodal LLMs perform on image classification of Christian saints? And \textbf{(RQ2)}, how does performance vary when enriching input with contextual information or few-shot exemplars?

We conducted a benchmarking study using three datasets supporting Iconclass natively: \textit{ArtDL}, \textit{ICONCLASS}, and \textit{Wikidata}, filtered to include the top 10 most frequent classes. Models were tested under three conditions: (1) classification using class labels, (2) classification with Iconclass descriptions, and (3) few-shot learning with five exemplars. Results were compared against ResNet50 baselines fine-tuned on the same datasets.

The findings show that \textit{Gemini-2.5 Pro} and \textit{GPT-4o} outperformed the ResNet50 baselines across the three configurations, reaching peaks of \textbf{90.45\%} and \textbf{88.20\%} in \textit{ArtDL}, respectively. Accuracy dropped significantly on the Wikidata dataset, where \textit{siglip-so400m-path-14-384} reached the highest accuracy score of \textbf{64.86\%}, suggesting model sensitivity to image size and metadata alignment. Enriching prompts with class descriptions generally improved zero-shot performance, while few-shot learning produced lower results, with only occasional and minimal increments in accuracy.

We conclude that general-purpose multimodal LLMs are capable of classification in visually complex cultural heritage domains, even without specific fine-tuning. However, their performance is influenced by dataset consistency and the design of the prompts. These results support the application of LLMs as metadata curation tools in digital humanities workflows, suggesting future research on prompt optimization and the expansion of the study to other classification strategies and models.
\end{abstract}


\section{Introduction}
For over two decades, the GLAM sector (galleries, libraries, archives, and museums) has undergone an extensive mass digitization process, resulting in a vast amount of digital archives containing a diverse array of artworks, photographs, and documents \cite{hawkins_archives_2022}. This rapid growth is transforming the analogical Cultural Heritage into a body of machine-readable knowledge, defining a critical mass of images and their metadata and serving as input to research in Artificial Intelligence (AI) and Computer Vision \cite{mishra_artificial_2024}. 

Computer Vision is a part of AI and Computer Science that enables computers to analyze, interpret, and make decisions based on image characteristics. At its core, it extracts and analyzes visual patterns, with applications that span the fields of medicine, robotics, and cultural heritage \cite{cetinic_understanding_2022, castellano_deep_2021}, with main tasks that include image classification \cite{cosovic_cnn_2020, jankovic_machine_2019, yu_image_2024}, object detection \cite{Gonthier_2018_ECCV_Workshops, inoue_2018_cvpr}, and semantic segmentation \cite{liu_recent_2019}. Among these, \textbf{image classification} stands out for its applicability to the semantically and visually complex objects in the Cultural Heritage field. For this task, the previously mentioned corpora of digitized artworks are well-suited for image classification, as they also contain metadata, usually created by domain experts, that describes the general content of the images. Of the several datasets, the ones with most extensive online resources are \textit{Art 500K} \cite{mao_deepart_2017}, the \textit{MET dataset} \cite{ypsilantis_met_2021}, and \textit{OMNIArt} \cite{strezoski_omniart_2018}, respectively, 500K, 400K, and 2M images of artworks labeled with attributes such as artist, genre, art movements, and materials. 

Another influential area of study supported with Image classification is Iconography, being that \textit{"branch of the history of art which concerns itself with the subject matter or meaning of works of art, as opposed to their form"} \cite{panofsky_studies_2018}. Iconography helps to understand the representations and themes expressed in images by identifying symbols, subjects, and motifs in paintings. Iconclass \cite{couprie_iconclass_1983} is a formal tool that can be utilized to support this area of study, offering a thesaurus that catalogs subjects and objects in artworks, including elements of historical, religious, and architectural significance. Thanks to this tool, image classification can be used to understand the overarching theme represented in an artwork, beyond surface-level object detection, without focusing on individual elements. Iconclass has been adopted as the backbone of other datasets, providing key metadata on specific visual features and serving as a foundation for the study of symbolic iconography. \textit{ArtDL} \cite{milani_dataset_2021} comprises approximately 42,500 images depicting Christian Saints, and \textit{DEArt} \cite{reshetnikov_deart_2022} contains more than 15,000 images featuring metadata, such as poses, with manually designated regions for the object detection task. The \textit{Iconclass AI Test Set} \cite{posthumus_iconclass_2020} deepens the focus by providing 87,500 images annotated using the complete set of Iconclass classes. \textit{IconArt} \cite{Gonthier_2018_ECCV_Workshops} features nearly 6,000 images tailored for object detection of Christian Saints and religious themes, such as the Crucifixion of Jesus and nudity.  Another important dataset is \textit{IICONGRAPH} \cite{sartini_iicongraph_2024}, which includes metadata for images from Wikidata and ArCo \cite{carriero_arco_2019}, modeled after the ICON Ontology \cite{sartini_icon_2023}, expanding the granularity to the context of Iconology \cite{panofsky_studies_2018}, and allowing cross-analyses of images based on the themes they represent.

These datasets have become more significant and widely utilized due to the substantial surge in the field of Computer Vision, where technical advancements enabled the use of Convolutional Neural Networks (CNNs) \cite{oshea_introduction_2015} and, more recently architectures centered on Vision Transformers (ViTs) \cite{dosovitskiy_image_2020}, enabling models to capture global dependencies in images rather than focusing on local features. To this family belong \textit{CLIP} \cite{radford_learning_2021} and \textit{SigLIP} \cite{zhai_sigmoid_2023}. They both align images and text in a shared embedding space through contrastive learning, excelling in zero-shot image classification. Beyond this, ViTs support multimodal Large Language Models (LLMs), including OpenAI \textit{GPT} \cite{radford_improving_2018}, Google \textit{Gemini} \cite{team_gemini_2023}, \textit{Claude} \cite{anthropic_claude_2024}, \textit{LLaVa} \cite{liu_visual_2023}, and \textit{Mistral} \cite{jiang_mistral_2023}. General-purpose systems capable of interpreting texts and images together, connecting complex visual and linguistic information. These models can be directly interfaced with full-text queries and analyze pictures, pushing the boundaries of automatic study as they can be adapted to a wide range of tasks without requiring retraining from scratch.

Over the last few years, several initiatives have been launched to leverage multimodal models and achieve state-of-the-art results in image classification. An example is fine-grained food image recognition through Vision Language Models (VLMs) \cite{kim_multimodal_2024}, leveraging \textit{CLIP} and image descriptions generated with \textit{MiniGPT-4} over data from two different datasets. Another study explores the use of off-the-shelf Multimodal LLMs, including \textit{CLIP}, \textit{SigLIP}, and \textit{BLIP-2} \cite{li_blip-2_2023}, in a zero-shot environment for classifying historical photographs in the Estonian Ajapaik archive \cite{maksimova_viability_2024}. The authors found these models to be underperforming in comparison to a fine-tuned supervised baseline for ambiguous or culturally specific categories, such as “viewpoint elevation”. For the case of image classification of Christian saints, the literature contains applications of CNNs to classify Christian iconography in artworks. For Example, in the paper introducing \textit{ArtDL} \cite{milani_dataset_2021}, the authors trained a \textit{ResNet50} model, achieving an accuracy of \textbf{84.44\%} in identifying the depicted saints. Other experiments yielded higher zero-shot accuracy in image classification by incorporating contextual information into the prompt. Only by describing sizes or media with prompts such as \textit{“a photo of a big \{class\}”} or \textit{“a photo of a small \{class\}”} leads to an additional \textbf{3.5\%} boost in accuracy with \textit{CLIP-ViT-L/14} \cite{zheng_large_2024} and also, a previously mentioned experiment, provided insight into enhancing accuracy results by implementing a framework that utilizes \textit{LLaMa2} to generate descriptions of images and enhancing the prompts for CLIP models, resulting in an approximately \textbf{9.6\%} increase in accuracy \cite{kim_multimodal_2024}.

These experiments demonstrate a growing interest in using multimodal models for general-purpose image classification tasks. Many advancements in the state-of-the-art have been achieved by implementing these models, and most importantly, by enhancing the accuracy scores of these models through the addition of meaningful context to the prompts \cite{kim_multimodal_2024}. Despite this interest, the literature has not yet benchmarked LLMs and VLMs specifically for classification in Christian iconography. This gap in the literature highlights the novelty of our preliminary investigation, which aims to address the following research questions: \textbf{(RQ1)} How do these models perform on the classification of Christian saints? \textbf{(RQ2)} How do the results change with the progressive enrichment of contextual data, such as adding more descriptive data or tagged exemplars?

In the following sections, we describe the methodology followed, highlighting the rationale behind the choice of the datasets and models. We then describe the technical implementations that can be found online in the published GitHub repository. The section is followed by the presentation of the results and their discussion, compared to the literature.

\section{Methodology}

In this analysis, we designed a study to evaluate the classification performance of diverse LLM models on Christian Iconography. Specifically, we aimed to (RQ1) compare different model architectures (VLMs and LLMs) in image classification and (RQ2) assess the impact of progressively enriching the input data with descriptions and few-shot exemplars. This section outlines the procedure design, including dataset preparation, model selection, and benchmarking.

\subsection{Datasets}

We investigated the images belonging to three collections, which we selected for their native support of Iconclass classes. One is \textit{ArtDL} \cite{milani_dataset_2021} and the other two are subsets of the \textit{ICONCLASS Test AI set} \cite{posthumus_iconclass_2020} and \textit{Wikidata}, of which we only include images representing the top ten most frequent classes of Christian Saints.

\subsubsection{ArtDL}

The \textit{ArtDL} dataset comprises over 42,000 images of Christian religious paintings. Each image is annotated with Iconclass codes, covering 10 key figures of Christian iconography, such as the Virgin Mary, Saint Francis of Assisi, and Saint Sebastian. The test set, published along with the paper, contains \textbf{1,864 images}. Each is mapped to a single iconographic label and serves as the starting point for our classification experiments presented in this study.

\subsubsection{ICONCLASS AI Test Set}

The \textit{ICONCLASS} \cite{posthumus_iconclass_2020} dataset originates from the official AI Test Set. The original collection comprises approximately 87,500 images representing a wide range of iconographic subjects. For this study, we conducted a filtering and curation process explicitly tailored to the single-label classification of Christian saints. Our refinement began by selecting all images annotated with Iconclass codes, starting with \textit{"11F"} (The Virgin Mary), \textit{"11H"} (male saints), and \textit{"11HH"} (female saints). We then performed a frequency analysis to identify the 10 most common saint classes. We applied systematic controls by removing all images with multiple saints, resulting in a dataset comprising \textbf{863 images}.

\subsubsection{Wikidata}

To expand our benchmark beyond curated archives, we compiled a dataset of religious artworks using the \textit{Wikidata SPARQL endpoint}. We designed a SPARQL query to detect paintings\footnote{In Wikidata are instances of type \textit{“wd:Q3305213”}} with a valid image URL and associated with Iconclass codes representing Christian saints and the Virgin Mary (using the same codes as the \textit{ICONCLASS} dataset). After filtering out multi-label entries, failed downloads, and duplicates, we retained \textbf{718 images} of paintings. Images were retrieved in their original size, preserving native aspect ratios. 

\subsubsection{Cross-Dataset Similarity Analysis}

\begin{figure}[t!]
  \centering
  \includegraphics[width=0.8\linewidth]{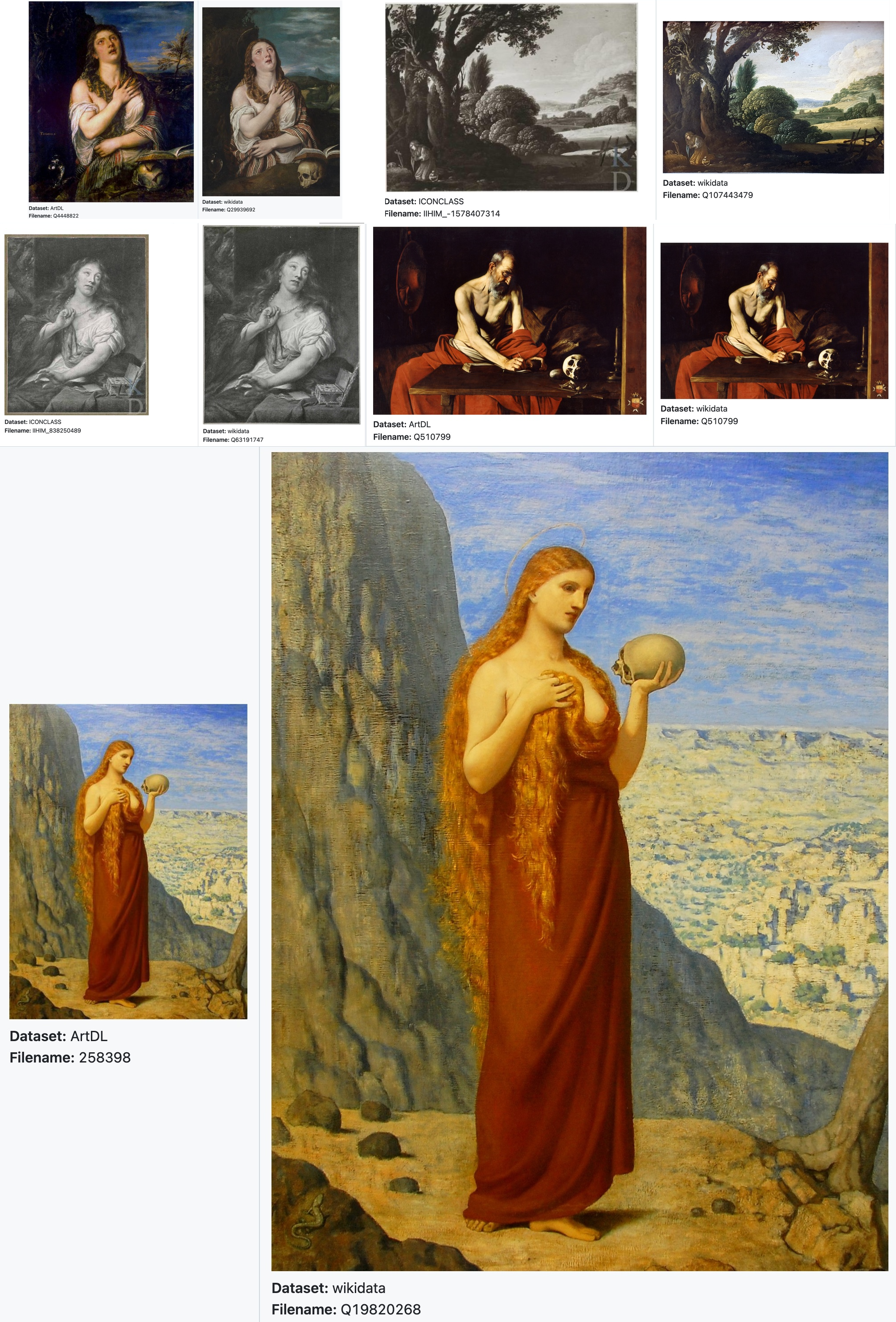}
  \caption{Example of image pairs deemed as similar (Hamming distance less than or equal to 8)}
  \label{fig:cross-dataset}
\end{figure}

To assess dataset independence and uncover overlaps that could bias evaluation results, we conducted a cross-dataset similarity analysis using a robust block-based image hashing method inspired by digital image forensics \cite{steinebach_robust_2012}. This approach identifies near-duplicate images across datasets, even those that have been transformed, such as cropping or mirroring. This method is particularly suitable for comparing art datasets, where visual duplicates may originate from different digitization pipelines or cataloging standards, resulting in varying representations of the same artwork. We defined a duplicate as any pair of images across datasets with a Hamming distance of 8 or less, identifying \textbf{36 cross-dataset duplicate} pairs using the robust hash, primarily between ArtDL and Wikidata (Figure \ref{fig:cross-dataset}).

\subsection{Models}

The study compares Vision-Language and Multimodal Large Language Models, adopting the Dual Encoder and Unified Transformer architectures, respectively, to highlight the differences in paradigms when processing single-labeled images of paintings depicting Christian Saints. The inclusion of these models reflects differences in the scale of parameters, input resolution, and the processing and presentation of visual and textual information.

For Dual Encoder architectures, we evaluated several variants of CLIP and SigLIP. CLIP models include the \textit{clip-vit-base-patch32} and \textit{clip-vit-base-patch16}, both with 86 million parameters and differing only in patch resolution, as well as the \textit{clip-vit-large-patch14} model, which significantly increases the model size to 304 million parameters with a finer 14$\times$14 patch granularity. Similarly, the SigLIP family includes \textit{siglip-base-patch16-512} and \textit{siglip-large-patch16-384}, differentiated by input resolution and internal capacity. The \textit{siglip-so400m-patch14-384} model is a larger variant with approximately 400 million parameters, trained on a broader corpus and optimized for robustness at an input size of 384$\times$384. These models encode images and text into a shared embedding space, relying on contrastive learning to align their representations, making them highly efficient for zero-shot classification.

In addition to contrastive models, we assess unified multimodal models, such as OpenAI \textit{GPT-4o (snapshot 2024-08-06)}, Google \textit{Gemini 2.5 Pro (preview-05-0)} classification smaller counterparts \textit{GPT-4o-mini (snapshot 2024-07-18)} and \textit{Gemini 2.5 Flash (preview-05-20)}. Flash and Mini are lightweight versions designed for efficiency and reduced cost while maintaining core multimodal capabilities. These models integrate both modalities into a single processing stream, enabling image understanding and contextual visual reasoning through prompt-based classification with natural language. They are particularly effective for visual studies, description generation, and interpretive tasks that require deeper semantic integration across modalities.

\subsection{Benchmarking design}

For building the benchmarking, we employed three distinct test configurations applied across models and datasets. This framework assesses model performance under varying conditions of knowledge availability. 

\paragraph{Test 1} is a zero-shot classification task with labeled names. The models classify images using only class label names (e.g., "St. Paul" or "Mary Magdalene") without additional contextual information. This configuration tests the model's ability to leverage pre-trained knowledge connections between visual features and semantic concepts. 

\paragraph{Test 2} is a zero-shot classification task with label descriptions. The Models receive detailed iconographic descriptions for each class, retrieved from Iconclass descriptions (e.g., "The penitent harlot Mary Magdalene; possible attributes: book (or scroll), crown …"). This approach evaluates how effectively models can utilize rich textual descriptions to guide visual classification. 

\paragraph{Test 3} is a few-shot Learning classification task. During inference, LLMs are provided with an arbitrary number of five example images, along with their corresponding class labels. For VLMs, we opted to fine-tune the last transformer layer using the same exemplars. The images have been chosen to represent the five least performing classes from the first test. This configuration assesses the models' capacity for rapid adaptation and in-context learning from limited examples.

\subsubsection{Technical implementation}

The benchmarking pipeline consists of a set of Python scripts available on the GitHub repository\footnote{https://anonymous.4open.science/r/Benchmarking-LLMs-for-Christian-Iconography-AA8F/README.md}, along with documentation that describes the detailed implementation points. At a high level, CLIP and SigLIP models were implemented using the HuggingFace interface, with a batch size of 16 images per classification. Resulting probabilities were computed using the model's native similarity output, e.g., Softmax for CLIP and Sigmoid activation for SigLIP. Both models output the probability distribution for each input image, enabling the evaluation of top-1 binary predictions. This means that in the resulting vector, the only class corresponding to the highest percentage was set to 1, while the others were set to 0.

GPT-4o was accessed through the OpenAI APIs and configured to behave deterministically with a \textbf{temperature value of 0} and an arbitrary \textbf{seed of 12345}, and forced to produce a JSON output. The evaluation batches contained five images, which have been proven to be less error-prone and more cost-effective than other configurations. Prompts followed a fixed-based template and dynamically included the list of class names or descriptions as candidate options\footnote{An example can be found on the GitHub repository linked above}. These outputs were parsed using a structured JSON extraction routine, with fallbacks for occasional formatting anomalies.

Similarly, \textit{Gemini 2.5} was tested using the Google AI Python SDK for the Gemini API. Likewise to GPT models, the \textbf{temperature was set to 0}, and the prompt was dynamically created by adding the list of candidate classes. These models have been configured with all safety categories (e.g., violent content, sexually explicit) set to \textit{BLOCK NONE}, ensuring that religious artworks are not automatically filtered out or rejected if they contain commonly found religious themes, such as nudity or martyrdom.

This study utilizes both open-source and proprietary multimodal models, each of which presents distinct technical implications. These closed-source models are accessed via API endpoints and have transparency limitations in their training data, internal representations, and filtering mechanisms, of which we only know external hyperparameters such as tokens and context windows. To mitigate output variance and ensure consistency in classification, LLMs were configured to operate in a near-deterministic mode, exploiting specific hyperparameter values. While this reduces stochastic variation, it does not eliminate the nondeterminism of LLMs.

\section{Results}

This section presents the preliminary results of the benchmarking experiments, which are structured to assess the classification performance of the models for each test in a progressive manner. We begin by establishing a supervised baseline. To ensure the reliability of our evaluation, we then include a cross-dataset analysis, which serves as a consistency check and measures whether models consistently assign stable predictions to similar images across the datasets. We then report overall accuracy scores across all models, datasets, and test configurations for a fine-grained benchmarking on iconographic classification. 

\subsection{Baseline}

To evaluate the models’ performances, we established a baseline based on supervised CNNs using \textit{ResNet50} architecture and employing the methodology from the \textit{ArtDL} paper \cite{milani_dataset_2021}. We leveraged two models, which were explicitly fine-tuned on an 80\% split of the images detected for the \textit{ICONCLASS} and \textit{Wikidata} datasets, respectively. The remaining 20\% has been used for testing. On ICONCLASS, this approach achieved an accuracy of \textbf{40.46\%}, while on Wikidata, it reached an accuracy of \textbf{43.97\%}.

\subsection{Cross-dataset consistency}

We evaluate the robustness of model predictions by conducting a cross-dataset consistency analysis across \textbf{36 matched image pairs} in the three test scenarios. The goal was to measure the percentage of image pairs for which both images received identical model predictions. The results are shown in Table \ref{tab:model_consistency_full}. 

Contrastive models exhibited higher consistency than multimodal models. Among the CLIP variants, consistency ranged from \textbf{55.56\%} to \textbf{86.11\%}, with the highest value achieved by \textit{CLIP-ViT-L/14} on Test 2, where 31 out of 36 matched pairs received identical predictions. SigLIP models showed broader variability, with consistency values from \textbf{33.33\%} to \textbf{77.78\%}. The strongest performance was achieved by \textit{SigLIP-SO400M-Patch14-384}, also on Test 2.

LLMs demonstrated lower consistency and greater sensitivity to the test configuration, particularly in terms of image sizes. \textit{GPT-4o} ranged from \textbf{25.00\%} in the first test to \textbf{27.78\%} in the second and third tests. \textit{GPT-4o-mini} had slightly lower overall scores but followed a similar trend. \textit{Gemini-2.5-Flash} achieved consistency between \textbf{30.56\%} and \textbf{33.33\%}, while \textit{Gemini-2.5-Pro} maintained a stable performance of \textbf{33.33\%} across all tests.

\begin{table}[h]
  \centering 
  \begin{tabular}{lcccc}
    \toprule
    \textbf{Model} & \textbf{Test 1} & \textbf{Test 2} & \textbf{Test 3} & \textbf{Avg. Consistency} \\
    \midrule
    clip-vit-base-patch32 & 58.33\% & 75.00\% & 61.11\% & 64.81\% \\
    clip-vit-base-patch16 & \textbf{66.67\%} & 75.00\% & \textbf{72.22\%} & \textbf{71.30\%} \\
    clip-vit-large-patch14 & 55.56\% & \textbf{86.11\%} & \textbf{72.22\%} & \textbf{71.30\%} \\
    siglip-base-patch16-512 & 38.89\% & 63.89\% & 33.33\% & 45.37\% \\
    siglip-large-patch16-384 & 41.67\% & 55.56\% & 41.67\% & 46.97\% \\
    siglip-so400m-patch14-384 & 55.56\% & 77.78\% & 61.11\% & 64.82\% \\
    gpt-4o-2024-08-06 & 25.00\% & 27.78\% & 27.78\% & 26.85\% \\
    gpt-4o-mini-2024-07-18 & 22.22\% & 30.56\% & 25.00\% & 25.93\% \\
    gemini-2.5-flash-preview-05-20 & 30.56\% & 30.56\% & 33.33\% & 31.48\% \\
    gemini-2.5-pro-preview-05-06 & 33.33\% & 33.33\% & 33.33\% & 33.33\% \\
    \bottomrule
  \end{tabular}
  \caption{Consistency results across three tests for all models, with the highest values in bold for each column.}
  \label{tab:model_consistency_full}
\end{table}

\subsection{Classification performances}

The classification performances are summarized in the following tables, showcasing the model’s accuracy for the three tests: \textbf{(1)} is Zero-Shot with only labels, \textbf{(2)} is a Zero-Shot setting with Iconclass descriptions, and \textbf{(3)} is a Few-Shot approach with labels.

\textit{ArtDL} (Table \ref{tab:artdl_performances}) performances varied significantly across models and configurations, ranging from \textbf{16.15\%} for \textit{CLIP-ViT-B/32} in test 1 to a peak of \textbf{90.45\%} achieved by \textit{Gemini-2.5 Pro}. Multimodal language models consistently outperformed contrastive learning models, with accuracies ranging from \textbf{82.46\%} to \textbf{90.45\%} across all settings, outperforming the baseline in almost all configurations, except for \textit{GPT-4o-mini} in the first test, which achieved an accuracy close to the baseline. The baseline model achieved an accuracy of \textbf{84.44\%}, surpassing VLMs.

\begin{table}[h]
  \centering 
  \begin{tabular}{lccc}
    \toprule
    \textbf{Model} & \textbf{Test 1} & \textbf{Test 2} & \textbf{Test 3} \\
    \midrule
    \textit{clip-vit-base-patch32} & 16.15\% & 31.55\% & 21.41\% \\
    \textit{clip-vit-base-patch16} & 25.64\% & 28.70\% & 29.13\% \\
    \textit{clip-vit-large-patch14} & 30.58\% & 44.31\% & 31.71\% \\
    \textit{siglip-base-patch16-512} & 48.71\% & 68.19\% & 55.90\% \\
    \textit{siglip-large-patch16-384} & 54.45\% & 72.21\% & 53.49\% \\
    \textit{siglip-so400m-patch14-384} & 53.86\% & 70.55\% & 56.38\% \\
    \textit{gpt-4o-2024-08-06} & 86.00\% & 87.45\% & 86.48\% \\
    \textit{gpt-4o-mini-2024-07-18} & 82.46\% & 84.98\% & 84.60\% \\
    \textit{gemini-2.5-flash-preview-05-20} & 88.20\% & 87.02\% & 84.71\% \\
    \textit{gemini-2.5-pro-preview-05-06} & \textbf{90.45\%} & \textbf{90.18\%} & \textbf{86.59\%} \\
    \textit{Baseline} & 84.44\% & 84.44\% & 84.44\% \\
    \bottomrule
  \end{tabular}
  \caption{Accuracy scores across three tests for \textit{ArtDL} dataset. The highest value per column is highlighted in bold.}
  \label{tab:artdl_performances}
\end{table}

For the ICONCLASS dataset (Table \ref{tab:iconclass_performances}), \textit{Gemini-2.5 Pro} achieved the highest performance, with accuracy scores of \textbf{83.31\%}, \textbf{84.82\%}, and \textbf{84.49\%} across the three evaluation settings. Among the dual encoders, \textit{SigLIP-SO400M-Patch14-384} performed best, achieving \textbf{60.88\%} in the Few-Shot setting and exceeding \textbf{50\%} in the other two configurations, aligning itself with the performances of \textit{GPT-4o-mini}. CLIP models produced moderate results, with \textit{CLIP-ViT-L/14} reaching a maximum of \textbf{42.81\%} accuracy in the Few-Shot configuration. \textit{GPT-4o} showed stable performance across the settings, and the baseline, fine-tuned on the \textit{ICONCLASS} dataset, achieved an accuracy of \textbf{40.46\%}.

\begin{table}[h]
  \centering 
  \begin{tabular}{lccc}
    \toprule
    \textbf{Model} & \textbf{Test 1} & \textbf{Test 2} & \textbf{Test 3} \\
    \midrule
    \textit{clip-vit-base-patch32} & 24.74\% & 29.30\% & 29.82\% \\
    \textit{clip-vit-base-patch16} & 30.00\% & 27.37\% & 33.51\% \\
    \textit{clip-vit-large-patch14} & 40.00\% & 35.44\% & 42.81\% \\
    \textit{siglip-base-patch16-512} & 43.51\% & 33.33\% & 41.93\% \\
    \textit{siglip-large-patch16-384} & 48.95\% & 38.77\% & 49.30\% \\
    \textit{siglip-so400m-patch14-384} & 59.47\% & 53.16\% & 60.88\% \\
    \textit{gpt-4o-2024-08-06} & 75.32\% & 75.43\% & 73.46\% \\
    \textit{gpt-4o-mini-2024-07-18} & 55.74\% & 59.56\% & 55.50\% \\
    \textit{gemini-2.5-flash-preview-05-20} & 77.17\% & 77.75\% & 78.22\% \\
    \textit{gemini-2.5-pro-preview-05-06} & \textbf{83.31\%} & \textbf{84.82\%} & \textbf{84.59\%} \\
    \textit{Baseline} & 40.46\% & 40.46\% & 40.46\% \\
    \bottomrule
  \end{tabular}
  \caption{Accuracy scores across three tests for \textit{ICONCLASS} dataset. The highest value per column is highlighted in bold.}
  \label{tab:iconclass_performances}
\end{table}

The \textit{Wikidata} dataset (Table \ref{tab:wikidata_performances}) generally showed lower accuracy scores compared to the other datasets, with most models clustering around the \textbf{45-60\%} range. Among all models, \textit{SigLIP-SO400M-Patch14-384} achieved the highest performance, with \textbf{66.29\%}, \textbf{59.60\%}, and \textbf{64.86\%} in the first, second, and third tests, respectively. Other SigLIP variants also performed strongly, with \textit{SigLIP-L/16-384} reaching up to \textbf{61.17\%}. Similarly, CLIP models showed moderate performance, with \textit{CLIP-ViT-L/14} achieving \textbf{56.76\%} in the zero-shot labels setting and slightly lower scores in the other configurations. In contrast to their strong results on previous datasets, \textit{GPT-4o} and \textit{Gemini-2.5} models demonstrated more modest performance here.\textit{GPT-4o} ranged between \textbf{45.31\%} and \textbf{45.75\%}, with \textit{GPT-4o-mini} falling behind. Likewise, \textit{Gemini-2.5 Pro} and \textit{Gemini-2.5 Flash} hovered around \textbf{44\%} to \textbf{47\%}, failing to match their high performance on other benchmarks. The baseline achieved \textbf{43.97\%} accuracy, outperforming \textit{GPT-4o-mini}, which highlights the difficulty of this dataset for large multimodal models.

\begin{table}[h]
  \centering 
  \begin{tabular}{lccc}
    \toprule
    \textbf{Model} & \textbf{Test 1} & \textbf{Test 2} & \textbf{Test 3} \\
    \midrule
    \textit{clip-vit-base-patch32} & 45.95\% & 44.52\% & 45.52\% \\
    \textit{clip-vit-base-patch16} & 50.78\% & 46.66\% & 47.08\% \\
    \textit{clip-vit-large-patch14} & 56.76\% & 56.61\% & 55.48\% \\
    \textit{siglip-base-patch16-512} & 57.47\% & 46.94\% & 56.05\% \\
    \textit{siglip-large-patch16-384} & 60.03\% & 43.95\% & 61.17\% \\
    \textit{siglip-so400m-patch14-384} & \textbf{66.29\%} & \textbf{59.60\%} & \textbf{64.86\%} \\
    \textit{gpt-4o-2024-08-06} & 45.75\% & 45.31\% & 45.31\% \\
    \textit{gpt-4o-mini-2024-07-18} & 35.78\% & 36.95\% & 34.31\% \\
    \textit{gemini-2.5-flash-preview-05-20} & 45.45\% & 45.31\% & 44.57\% \\
    \textit{gemini-2.5-pro-preview-05-06} & 45.89\% & 45.31\% & 47.07\% \\
    \textit{Baseline} & 43.97\% & 43.97\% & 43.97\% \\
    \bottomrule
  \end{tabular}
  \caption{Accuracy scores across three tests for \textit{Wikidata} dataset. The highest value per column is highlighted in bold.}
  \label{tab:wikidata_performances}
\end{table}

Across all three datasets, Large Language Models, particularly \textit{Gemini-2.5 Pro} achieved the highest accuracy scores, consistently outperforming other models on ArtDL and ICONCLASS. On \textit{Wikidata}, we also observed a general decline in accuracy. While top-performing models often exceeded \textbf{80\%} accuracy on \textit{ArtDL} and \textit{ICONCLASS}, performance on \textit{Wikidata} was relatively lower, reflecting a more challenging classification scenario. The \textit{ResNet50} baselines were often outperformed by LLMs, despite being fine-tuned directly on the target datasets.

Performance across the three evaluation configurations varied by dataset. On \textit{ArtDL}, accuracy generally improved across the three stages, with few-shot learning leading to lower results, with only two cases, \textit{Gemini 2.5 Flash} on \textit{ICONCLASS} and \textit{Gemini 2.5 Pro} on \textit{wikidata} reaching \textbf{1-2\%} increment in accuracy. On \textit{ICONCLASS}, the pattern was less uniform but still showed improvement in several cases. On \textit{Wikidata}, performance remained relatively stable across the three configurations, with no consistent trend of improvement.

\section{Discussion}

In this section, we discuss the performance of different model architectures and prompt designs in the task of classifying Christian iconography (\textbf{RQ1}) and how the results change after the progressive enrichment of contextual data (\textbf{RQ2}).

For \textbf{RQ1}, the results show that multimodal LLMs generally outperform traditional supervised models for the task of image classification of Christian iconography. For \textit{ArtDL} and \textit{ICONCLASS} datasets, \textit{Gemini-2.5 Pro} achieved the highest accuracy in all three evaluation settings, surpassing fine-tuned \textit{ResNet50} baselines. These findings confirm the effectiveness of multimodal LMMs in semantically dense and specific tasks. The Classification for \textit{Wikidata} images witnessed overall lower accuracies, with the best performance from \textit{SigLIP-SO400M-Patch14-384}, achieving \textbf{66.29\%} accuracy in zero-shot classification and surpassing both Gemini and GPT-4o. This behavior suggests that contrastive architectures tend to perform better when facing inconsistent image qualities and sizes, as confirmed also by the consistency check, where the same image is predicted differently when the sizes differ. 

Additionally, the studied LLMs are likely being trained on ArtDL and ICONCLASS datasets, as it is common knowledge that the publishers scrape the Internet for training data. At the same time, the full Wikidata knowledge base, which presents denser metadata, may not have its specific weights corresponding to the Iconclass codes tuned efficiently. Additionally, the two supervised \textit{ResNet50} baselines, despite being fine-tuned only on a small set of 600 images, demonstrate that recent multimodal LLMs outperform supervised approaches, even for ArtDL, which has a more consistent number of images.

\begin{figure}[t!]
  \centering
  \includegraphics[width=0.8\linewidth]{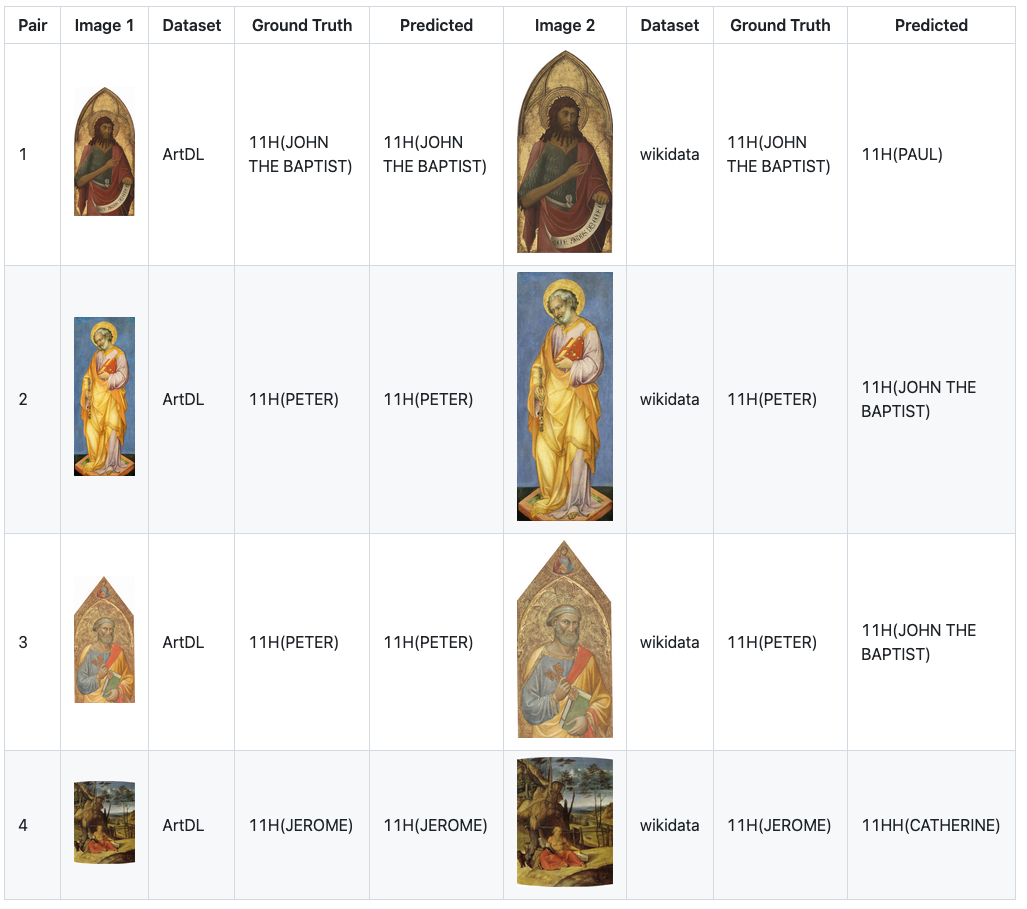}
  \caption{Example of incorrectly predicted images for Gemini 2.5 Pro, test 2. The image classification per row, one from ArtDL on the left and one from Wikidata. The focus is on the predicted class, differing for each dataset.}
  \label{fig:errors}
\end{figure}

Regarding \textbf{RQ2}, we found that zero-shot classification using label descriptions generally improved accuracy except for \textit{Wikidata}, where the changes are mostly negatives for LLMs, while generally improving accuracy for contrastive models. \textit{CLIP-ViT-L/14} on ICONCLASS increased from \textbf{40.00\%} to \textbf{42.81\%} with the addition of textual descriptions. This finding is in line with previous studies \cite{zheng_large_2024, kim_multimodal_2024} emphasizing the importance of injecting the prompts with semantic information describing the content. While \textit{Gemini 2.5 flash} in \textit{ICONCLASS} and \textit{Gemini 2.5 Pro} in \textit{Wikidata} demonstrated slight improvements in accuracy, the other language models performed worse. This outcome proves that few-shot learning does not always improve results and highlights the possible sensitivity to prompt formatting, class imbalance, or overfitting to a few visual characteristics. This also highlights the sensitivity to poorly chosen exemplars, which can degrade performance rather than enhance it.

Despite these results, several limitations should be acknowledged. This study focuses on classifying single-labeled images, and the number of classes per dataset was relatively low; these limitations do not reflect a real-world scenario. Additionally, few-shot prompting did not consistently improve results, suggesting that the task is non-trivial. Further research is needed to optimize prompt formatting and choose the correct few-shot exemplars.
These findings provide empirical support for using general-purpose multimodal models for the iconographical classification of artwork images. In particular, they validate the applicability of multimodal LLMs for low-data and high-complexity domains, such as Christian iconography, where semantic and label overlap is common. Even with these limitations, state-of-the-art models can "detect" and "classify" \textit{Saint George} or \textit{Saint Sebastian}, suggesting that these tools may help scholars in metadata curation without the need for extensive retraining.

\section{Conclusions}

This study serves to benchmark VLMs and multimodal LLMs for the task of classifying Christian iconography, a domain with limited training data. The output is a reproducible evaluation framework spanning three datasets and classification settings, providing a baseline for future research in applying general-purpose AI systems to cultural heritage tasks. The results demonstrate that \textit{Gemini 2.5 Pro} and \textit{GPT 4o} can outperform traditional supervised baselines, indicating their potential as tools for metadata enrichment and semantic indexing in Digital Humanities workflows \textbf{(RQ1)}. Prompt enrichment improved performance in most settings, but few-shot learning mostly led to lower outcomes \textbf{(RQ2)}, suggesting that a more optimal example selection is required. This could be addressed by integrating Retrieval Augmented Generation (RAG) pipelines while also reducing the risk of hallucinations. 
For future works, we aim to extend this benchmarking with real-world scenarios, where classification is required for polyptychs or paintings featuring multiple saints. We need to integrate various visual and textual elements through training on datasets specifically providing this information.

\section*{Acknowledgements}

The authors acknowledge Villa I Tatti, the Harvard University Center for Italian Renaissance Studies, for providing access to a dedicated computation server equipped with a GPU, which was essential for running vision-language model inference and fine-tuning. Villa I Tatti also provided API access to the GPT and Gemini 2.5 models, enabling the experiments presented in this work.

\bibliographystyle{plain}
\bibliography{bibliography}

\end{document}